# Can Foundational Large Language Models Assist with Conducting Pharmaceuticals Manufacturing Investigations?

Hossein Salami[1*], Brandye Smith-Goettler[2], Vijay Yadav[2]

[1] Digital Services, MMD, Merck & Co., Inc., 126 E. Lincoln Ave., Rahway, NJ 07065, USA

[2] Digital Services, MMD, Merck & Co., Inc., 770 Sumneytown Pike, West Point, PA 19486, USA

**Abstract**

General purpose Large Language Models (LLM) such as the Generative Pretrained Transformer (GPT) and Large Language Model Meta AI (LLaMA) have attracted much attention in recent years. There is strong evidence that these models can perform remarkably well in various natural language processing tasks. However, how to leverage them to approach domain-specific use cases and drive value remains an open question. In this work, we focus on a specific use case, pharmaceutical manufacturing investigations, and propose that leveraging historical records of manufacturing incidents and deviations in an organization can be beneficial for addressing and closing new cases, or de-risking new manufacturing campaigns. Using a small but diverse dataset of real manufacturing deviations selected from different product lines, we evaluate and quantify the power of three general purpose LLMs (GPT-3.5, GPT-4, and Claude-2) in performing tasks related to the above goal. In particular, (1) the ability of LLMs in automating the process of extracting specific information such as root cause of a case from unstructured data, as well as (2) the possibility of identifying similar or related deviations by performing semantic search on the database of historical records are examined. While our results point to the high accuracy of GPT-4 and Claude-2 in the information extraction task, we discuss cases of complex interplay between the apparent reasoning and hallucination behavior of LLMs as a risk factor. Furthermore, we show that semantic search on vector embedding of deviation descriptions can be used to identify similar records, such as those with a similar type of defect, with a high level of accuracy. We discuss further improvements to enhance the accuracy of similar record identification.

# 1. Introduction

In a manufacturing environment, incidents refer to any unforeseen event occurring in a product manufacturing line. These incidents can be deviations from specific operational plans, protocols, standard operating procedures, guidelines, etc., and are important as they potentially can result in production halts, safety or environmental issues, or product quality concerns. These side effects can lead to increased cost, reduced productivity, or product loss which makes the identification, resolution, and prevention of incidents and deviations to be critical in any manufacturing environment. Depending on the significance and specific requirements, a deviation occurrence might trigger a systematic and comprehensive investigation process. In general, such investigation aims at identifying main underlying causes leading to the incident, potential impacts on product and quality, and framing preventive actions to minimize the probability of similar occurrences in the future. Performing a full investigation can take days to weeks and might require an interdisciplinary team of experts from engineering, operations, quality, safety, data science and other groups.

In many cases, these investigations do not leverage the prior knowledge from similar cases which makes conducting them even more time and resource consuming. A typical manufacturing organization might have faced hundreds of incidents and deviations in the past with new ones being recorded every year. A majority of incidents are likely to be simple deviations with straightforward root cause identification and a minimal impact. Nevertheless, a growing body of knowledge from historical cases exists that might be helpful in dealing with future cases. The acquired knowledge from handling previous incidents is mostly stored in two forms. Personnel in engineering, quality or other related groups are likely to gain significant experience in dealing with certain categories of incidents, appropriate actions to take to minimize the associated risks, or where to look first to identify potential root causes. Besides this tactic knowledge, most organizations would have a defined framework to formally record incidents and deviations. In more regulated sectors such as pharmaceutical manufacturing, quality investigation reports are prepared, reviewed, and stored to be available during possible inspections. A previously investigated deviation might help in hypothesizing potential causes for a new case with a somewhat similar description, taking corrective actions to deal with it, or defining preventive actions for a specific product line or a new manufacturing campaign. Naturally, new cases might involve completely new causes and require new treatments. However, incidents with common or at least somewhat similar themes, root causes, corrective actions, etc. do exist. Therefore, it stands to reason to hypothesize that some incident records and lessons learned from them might enable one to respond to a new incident in a shorter time, and in a more efficient and effective manner. Also, in a more holistic picture, information from these records can enable an organization to find trends and common themes in incidents or take steps to minimize risks in new campaigns.

For a mature manufacturing organization, it is unreasonable to expect any single member to have a full knowledge of all the investigation records and their details. Automated approaches enabled by IT capabilities, databases, and AI and data analytics tools are necessary. Specifically for recorded investigation reports, as a mainly unstructured source of data, one might think of leveraging Natural Language Processing (NLP) techniques for automation. The NLP field started on a transformational journey in 2017 upon publication of the *Transformer* architecture, a neural network-based model capable of processing sequence data such as natural language and effectively capturing long-range dependencies (1). More recently, the introduction of general-purpose Large Language Models (LLM) such as Large Language Model Meta AI (LLaMA) (2) and Generative Pre-trained Transformer (GPT) (3) has led to a significant rise in interest from different sectors, and a variety of proposed applications.

Despite many ongoing efforts, application of LLMs in industries and settings with high accuracy standards is complicated by several factors. Unlike a typical NLP study, in most proposed use cases there are no task and problem-specific benchmarks to evaluate an LLM's performance. At least for now, even a clear problem definition and success metric might not exist for some loosely proposed applications. These factors combined with concerns such as quality or availability of data, hallucinations, regulations, biases, etc. pose serious questions and barriers on the path to the full realization of LLMs' potential benefits in some industries.

In this work, we aim at evaluating the potential application of general-purpose LLMs in assisting with conducting manufacturing investigations. In approaching this question, we are cognizant that these are mathematical models, and not intelligent agents that excel at all tasks on all data. Therefore, we break down the problem into specific subsets, each involving a certain NLP-related task that can be useful in the context of manufacturing investigations; we provide quantified results on the performance of example LLMs for each task. In the absence of public benchmarks, for different tasks we define the ground truth by either using existing records and metadata or by manually inspecting documents and scoring model outcomes.

## 2. Dataset and model development

To evaluate the ability of example general-purpose LLMs in performing different tasks related to manufacturing investigations, a small dataset containing 20 instances was constructed. Each instance corresponds to a full investigation report prepared in response to a real pharmaceutical manufacturing deviation. Each report is also accompanied by a set of metadata such as date of the occurrence that were recorded separately. A compiled investigation report typically contains several information including details of the deviation, actions taken in response to the discovery of deviation, identified root causes, corrective and preventive actions, and an assessment of the possible impact on quality. The only pre-processing step that was applied to the extracted text from each report was removal of extra spaces and lower casing. Throughout the manuscript phrases incident, deviation, or investigation all refer to instances in the dataset.

Overall, three foundational LLMs were evaluated for different tasks. OpenAI's GPT-3.5-Turbo-16k-0613 and GPT-4-32k-0613 models along with Anthropic's Claude-2 were tested as sequence-to-sequence models (i.e., chat-like conversations where a model accepts a string sequence as input and returns a new string in response), and OpenAI's ada-002 model was tested for text embedding generation (more details about embedding generation is provided in the next section). These models were selected from a fast-growing body of available options. OpenAI's GPT series have shown to perform well on various tasks including GPT-4 scoring in the 89[th] percentile in SAT math (4), Anthropic's Claude-2 reportedly scored above the 90[th] percentile on the graduate record examination (GRE), highlighting capabilities of both models (5). We note that similar analysis can be performed for other available options. New and upcoming models are likely to be even more powerful than the current ones. In all cases the general-purpose models were used and no generative pre-training or supervised fine-tuning was performed (for examples of such look at (3)). All models were accessed by making API calls to the related hosted endpoints.

## 3. Results and discussion

**Zero-shot learning**

An obvious use for a general-purpose LLM in any context is to use it as a zero-shot learner. Zero-shot learning (Figure 1) refers to cases in which a question is presented to an LLM as an input sequence, without providing any further context and supplementary information (i.e., in-context learning (6, 7)), or any example of the desired output (i.e., few-shot learning (6, 8), here the phrase *learning* should not be confused with learning as the process of training a model and adjusting its internal parameters). There are reports of successful uses of models like GPT in this manner, obtaining high performance scores on certain datasets. (3, 4, 9) In this framework the only source of information is model parameters (e.g., neural network weights) in which certain world knowledge is implicitly stored (i.e., parametric memory) (9, 10). This approach suffers from clear drawbacks including requiring large models with large number of parameters to store more knowledge, and the need for constant fine-tuning in order to have access to the most up-to-date or domain-specific information. (9)

As mentioned, already some of the published LLMs show superb zero-shot performance in certain tasks. However, it is unlikely that using general-purpose LLMs as purely zero-shot learners can provide a significant benefit to manufacturing investigations. Most useful questions would be too specific to an organization, to a manufacturing process step, etc. A language model trained on public data would not have access to such information during its training and so would be unable to provide useful answers. Nevertheless, for more general inquires, such as common technical definitions, or questions dealing with common engineering knowledge these models might be helpful, but still subject to common concerns such as hallucination.

As a brief side note, it should be noted that training a language model on public data still can be beneficial as it enables the model to learn various structures of the natural language. For example, it has been shown that an unsupervised generative pre-training before a task-specific fine-tuning step can improve the performance of LLMs (3).

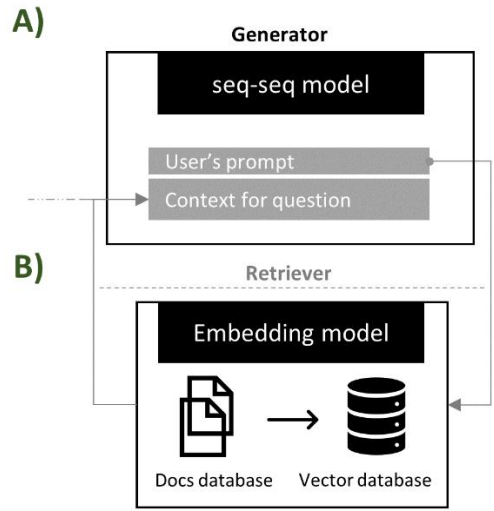

**Fig. 1.** An example architecture for using a language model to assist with investigations. (A) Using a sequence-to-sequence model such as GPT-3.5 or LLaMA to generate a response to an input text. (B) Providing context for user's input by, for example, retrieving related documents from an external knowledge base before generating a more accurate and relevant response, using similarity between embeddings that are produced using a text embedding model. The retriever can also take advantage of other search mechanisms such as keyword-based search. Zero-shot inference or learning refers to cases in which the user's prompt is the only input to the model without any additional context.

**Extraction of incident details**

A simple but illustrative use case related to manufacturing investigations is leveraging an LLM to extract various details from an investigation report. Access to this information enables organizations to have a more holistic view of the frequency of incidents with certain natures, detect common themes, potentially identify manufacturing steps with highest probability of certain incidents, etc. Here we limit the task to the extraction of five pieces of information from each report: Date and site of occurrence, involved product batches, whether the incident had any quality impact, and the identified root cause for the event. While there might be other details that are of interest, the benefit of limiting the task to this information is that a simple metric can be used to quantify the performance of each model. Accordingly, for each LLM we count the number of reports from the dataset for which accurate information are extracted. For each instance, the input to the LLM consists of 1- an introductory prompt, 2- a context for the problem, which is the full text from the report associated with that deviation, and 3- a question prompting the model to return the requested information using the provided context (Figure 1).

Figure 2 shows the performance of each model in extracting the requested information from the investigation texts. Each response is scored to be either Accurate, Acceptable, or Inaccurate. The Acceptable category refers to cases in which either a single well-defined ground truth value could not be established, or when the model outcome failed to capture some of the minor details. All models score full on one out of the five tasks (date of occurrence). On the other four tasks Claude-2 and GPT-4 show superior performance compared to GPT-3.5 which was somewhat expected based on our initial tests and publicly available discussions and data. In the task of extracting the identified root cause for a deviation, a task which requires stronger logic, GPT-4 performs slightly better than Claude-2. For both models, however, in several cases one of the major causal factors in the process was identified as the main root cause. These are, however, marked as Acceptable since they can be mostly resolved with prompt optimization.

Another intriguing observation related to the data in Figure 2 was an inaccurate response from Claude-2 regarding the manufacturing site of a deviation. In the text associated with that specific instance, there was no mention of the manufacturing site, however, the text contained information about a product made by Company-A, and also mentioned the name of an independent contract manufacturing organization (CMO). Interestingly, in the model's response the manufacturing site belonging to Company-A and located in the same State-X as in the CMO, was reported as the manufacturing site associated with that incident. This is an interesting case since it highlights both a strong ability by the model for logical interpretation (7), as well as the complexity of forms in which an LLM can hallucinate (the incident report is from Company-A, Company-A has many sites including one in State-X, the report also mentions a CMO which is located in State-X, therefore, the incident site should have been Company-A's site in State-X. Indeed, this was supported by model's response when the model was prompted to explain the logic of arriving at its answer).

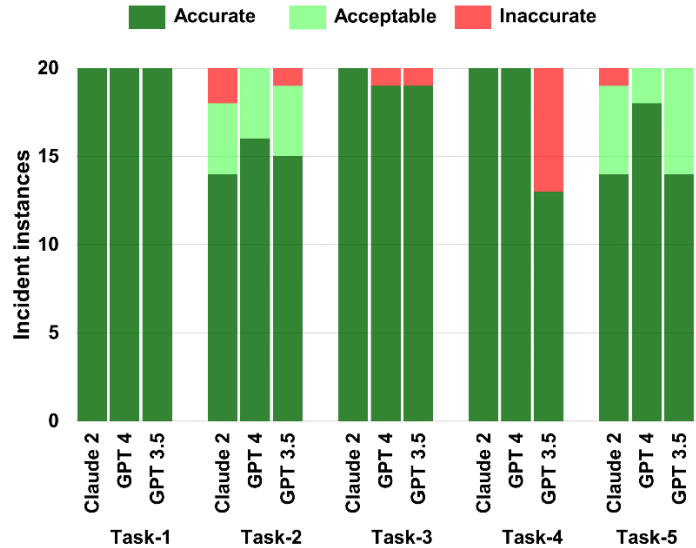

**Fig. 2.** Comparing the performance of different LLMs on example tasks related to analyzing manufacturing investigation reports. Task-1: Identification of the date of occurrence, Task-2: Identification of the manufacturing site, Task-3: Identification of the impacted product batch, Task-4: Whether the incident led to quality impact on the batch or not, Task-5: Identified root cause for the incident. Acceptable category refers to cases in which either a single well-defined ground truth value could not be established, or when model outcome failed to capture some of the minor details.

**Text relatedness**

One opportunity of having access to records of previous deviations in a manufacturing environment is to use the knowledge gathered from those in addressing and investigating new cases. To achieve this goal, a key step is to identify records most relevant to a new case under study. For example, facing an incident involving broken glass vials one can retrieve reports containing words "glass", "broken", "vial" and potentially leverage those as prior related knowledge and experience. There are clear drawbacks with this approach however, including the fact that these *keywords* by themselves might not communicate the full story behind an incident and its attributes such as root cause. Therefore, this somewhat defective search of historical records might lead to missing relevant and useful documents or retrieval of documents that are not so useful.

An alternative way is to retrieve documents based on the similarity of their vector representations with that of a reference text such as description of the new incident under study. Vector representations, or text embeddings, can be generated using a pre-trained encoder (10, 11). Such encoders are essentially mathematical transformations that map input words or sentences to a vector space where, ideally, sentences with similar or related meaning (as a simple example, consider a paragraph and its paraphrase) are close to each other. (12) This closeness or relatedness (i.e., semantic similarity) can be quantified by vector similarity metrics such as cosine similarity or L2 distance). This vector-based search is one underlying idea behind the much talked about Retrieval-Augmented-Generation (RAG) methodology, in which a pre-trained LLM can be used to respond to domain-specific inputs by leveraging an external knowledgebase (Figure 1). A retriever finds the most *relevant* documents from a knowledgebase and provides them to a generator model, as context, along with user's input such as a question (for more detailed description and examples of different RAG methodologies look at (8-10, 13)).

Here, we evaluate an example text embedding model, OpenAI's text-embedding-ada-002 for identifying related manufacturing deviations in our dataset. Accordingly, for each instance in the dataset a short paragraph describing the incident was prepared using the reference investigation report and descriptions in there. Each paragraph was then vectorized using the text embedding model to generate a fixed-length (1536, 1) embedding vector associated with that instance. Cosine similarity was used to calculate the similarity between each pair of deviations in our dataset (Figure 3).

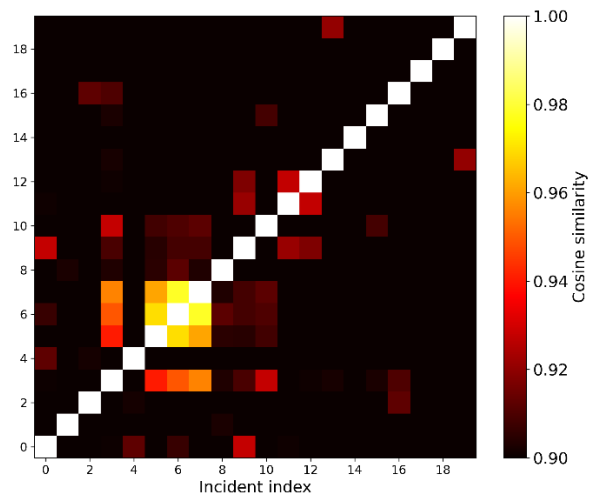

**Fig. 3.** Embedding vectors cosine similarity map for incident descriptions. Embedding vectors are generated using the OpenAI's text-embedding-ada-002 model, each embedding is a fixed-length (1536, 1) vector.

Analyzing Figure 3, all diagonal elements have value of 1 (i.e., completely similar vectors) which is expected as they correspond to identical incident descriptions. For several pairs including records (3, 5), (0, 9), (11, 12), and (6, 7) there is a very high similarity between incident descriptions based on the embedding vectors. We closely analyzed some of these cases by manually inspecting each pair and check whether the suggested relatedness between the two incidents (based on vector embeddings) is valid. Indeed, for all cases pairs with high similarity values have somewhat related descriptions. Table 1 shows some parts of the incident description for a couple of example pairs that have high similarity based on the data in Figure 3. For example, instances 3 and 5 in the dataset both describe incidents in which particle alarms were raised due to false rejects (although they had different root causes). Instances 0 and 9 both refer to incidents in which a particle defect in the form of visible particles was identified. Instances 13 and 19 both refer to incidents involving glass breakage in the manufacturing line. Note that these "related" cases do not necessarily share every detail such as the root cause, they simply have similar descriptions based on the vector embeddings generated by a certain text embedding model. Nevertheless, based on this limited data, it seems that vector representations generated by this model are useful for detecting related incidents.

**Table 1.** Example of incident descriptions with high cosine similarity between vector embeddings according to Figure 3. Provided descriptions are masked due to the proprietary nature of the dataset.

| **Records 3 and 5** |
|---|
| - During the inspection of … the particle rejects function surpassed its limit… creation of excess microbubbles … causing false rejects … |
| - During batch reconciliation … particle rejects process control limit … exceedance was due to false rejection as demonstrated by … |
| **Records 0 and 9** |
| - On … during the routine … in the form of a … particle … |
| - On … during a manual … was identified by … was categorized as a particle defect … |
| **Records 13 and 19** |
| - On … an alarm in … halting the process … be a piece of glass … |
| - On … broken … glass to glass contact … during … manufacturing process … |

Data presented in Figure 3 and Table 1 support using vector-based semantic search for identifying relevant and related incidents from historical records. As it was shown, at least for our small dataset, incidents with similar or related descriptions had close vectors in the embedding space as well. However, as shown in Figure 3, the similarity score between many other pairs in the dataset (although referring to mostly un-related incidents) is also relatively high (e.g., 0.9). This is somewhat expected as all the records in this dataset are texts related to manufacturing incidents and in the same industry, a very specific domain in the natural language landscape. This overall relatedness makes it challenging for a general-purpose trained embedding model, such as the one used here, to distinguish between these instances fully and clearly.

Depending on specifics of business and use case other top document selection procedures such as similarity based on root cause, or filtering documents based on associated metadata (e.g., limiting the search to incidents in a certain manufacturing product line), or simple keyword-based filters (e.g., limiting the search to documents with the phrase "broken vial") might be useful for improved document retrieval. The retrieved related cases might then be used as part of a RAG pipeline to potentially assist with conducting the new investigation.

## 4. Conclusion and outlook

In this work we evaluated the performance of example foundational LLMs in a few tasks related to leveraging historical records of manufacturing deviations and investigations to gain insights and reduce the time to deal with new cases. Using a small dataset of real manufacturing deviation reports, we evaluated the power of general purpose LLMs to extract information about a case based on the text of its report or finding similar or related cases using embedding vectors.

Large language models have generated much excitement in recent years. While there is strong evidence that these models can perform remarkably well in various NLP tasks, their application to domain-specific cases, including assisting with conducting manufacturing investigations requires additional steps to be taken. The first and most important step is to frame well-defined problem statements. This step is critical as different tasks might rely on different capabilities and types of language models. Take, for example, the problem of finding similar or related incidents in historical records. This as a standalone task mostly does not rely on "generative" capability of sequence-to-sequence models such as LLaMA or GPT, rather on producing semantically relevant embedding vectors using a text embedding model. Failure to frame specific problem statements and sub-tasks leads to slow progress, confusion, and bucketing everything and all components of a tool under development into a vaguely defined "Gen-AI" bucket. Well-defined tasks enable evaluating performance of different models and workflows more systematically and clearly. Certain language models or model architectures might perform better for specific domains or tasks. Considering the fast progress of the field and the likely emergence of new models and tools, it is important to be able to distinguish between various options.

As it has been widely reported and was also shown above, LLMs can show strong apparent reasoning capabilities. This is a powerful tool, however, as it was seen in one instance in our data, that can lead to model generating inaccurate responses. Preventions and guardrails such as having a human in the loop might be necessary specially in high-risk tasks. In many cases probability of inaccurate responses can be lowered by optimizing input prompts (for example, in the inaccurate site identification example above). However, prompt engineering is an evolving area of research, there are examples of successful approaches, such as chain-of-thought (14), however, not every prompt engineering method is applicable to all tasks. Furthermore, issues such as prompt instability (15) and non-existent ground truth data for some applications makes defining the "best" prompt challenging.

Depending on specifics of a business area there might be interest in leveraging model fine-tuning to improve performance of LLMs for certain purposes. This can be the case when alternate strategies such as RAG is either not successful or not applicable. Note that fine-tuning, in the machine learning sense of the word, refers to adjusting the internal parameters of a model to improves its accuracy regarding a certain task. This is a complicated step. Most importantly, preparing appropriate labeled data to use for supervised fine-tuning of a model can be very resource intensive. Supervised training of a general purpose LLM might have been based on labeled datasets including question-answer pairs. Unlike this very well-defined and straightforward to prepare data (although still time consuming), preparing a labeled dataset for fine-tuning a language model for a specific domain such as manufacturing investigations is not trivial at all. Also, one needs to ask if fine-tuning a model can add any benefit to their workflow or digital tool they are trying to build in the first place.

Further exploration into advantages and benefits that language models can provide in the manufacturing space should be accompanied by framing clear problem statements in which the language model-related components are decoupled from other elements. Gathering domain-specific data and benchmarks enable the evaluation and scoring of different models. Leveraging new research findings and more sophisticated methods such as novel prompt engineering methods, learned document retrieval, or multi-modal models that can leverage information in the form of both text and image should be considered for future steps.

## 5. References

1. Vaswani A, Shazeer N, Parmar N, Uszkoreit J, Jones L, Gomez AN, et al. Attention is all you need. Advances in neural information processing systems. 2017;30.
2. Touvron H, Lavril T, Izacard G, Martinet X, Lachaux M-A, Lacroix T, et al. Llama: Open and efficient foundation language models. arXiv preprint arXiv:230213971. 2023.
3. Radford A, Narasimhan K, Salimans T, Sutskever I. Improving language understanding by generative pre-training. 2018.
4. OpenAI [cited 2024 Feb 1]. Available from: [www.openai.com/research/gpt-4](www.openai.com/research/gpt-4).
5. Anthropic [cited 2024 Feb. 1]. Available from: [www.anthropic.com/news/claude-2](www.anthropic.com/news/claude-2).
6. Brown T, Mann B, Ryder N, Subbiah M, Kaplan JD, Dhariwal P, et al. Language models are few-shot learners. Advances in neural information processing systems. 2020;33:1877-901.
7. Saparov A, He H. Language models are greedy reasoners: A systematic formal analysis of chain-of-thought. arXiv preprint arXiv:221001240. 2022.
8. Izacard G, Lewis P, Lomeli M, Hosseini L, Petroni F, Schick T, et al. Few-shot learning with retrieval augmented language models. arXiv preprint arXiv:220803299. 2022.
9. Ram O, Levine Y, Dalmedigos I, Muhlgay D, Shashua A, Leyton-Brown K, et al. In-context retrieval-augmented language models. arXiv preprint arXiv:230200083. 2023.
10. Lewis P, Perez E, Piktus A, Petroni F, Karpukhin V, Goyal N, et al. Retrieval-augmented generation for knowledge-intensive nlp tasks. Advances in Neural Information Processing Systems. 2020;33:9459-74.
11. Devlin J, Chang M-W, Lee K, Toutanova K. Bert: Pre-training of deep bidirectional transformers for language understanding. arXiv preprint arXiv:181004805. 2018.
12. Chollet F. Deep learning with Python: Simon and Schuster; 2021.
13. Guu K, Lee K, Tung Z, Pasupat P, Chang M, editors. Retrieval augmented language model pre-training. International conference on machine learning; 2020: PMLR.
14. Wei J, Wang X, Schuurmans D, Bosma M, Xia F, Chi E, et al. Chain-of-thought prompting elicits reasoning in large language models. Advances in Neural Information Processing Systems. 2022;35:24824-37.
15. Liu X, Zheng Y, Du Z, Ding M, Qian Y, Yang Z, et al. GPT understands, too. AI Open. 2023.